\pdfoutput=1
\documentclass[journal]{IEEEtran}
\IEEEoverridecommandlockouts       

\usepackage{amssymb}
\usepackage{amsmath}
\usepackage[english]{babel}
\usepackage{float}
\usepackage{graphicx}
\usepackage{hyperref}
\usepackage{mathtools}

\usepackage{filecontents}
\usepackage[noadjust]{cite}
\usepackage[font=footnotesize,labelfont=footnotesize]{subcaption}
\usepackage[font=footnotesize,labelfont=footnotesize]{caption}
\usepackage{comment}
\usepackage{multirow}
\usepackage{xcolor}
\usepackage{algorithm}
\usepackage{algorithmic}
\usepackage{xspace}
\usepackage{graphicx}
\usepackage{booktabs}

\definecolor{darkblue}{rgb}{0.15,0.15,0.55}
\definecolor{lightgrey}{rgb}{0.75,0.75,0.75}

\providecommand{\codecomment}[1]{\textcolor{lightgrey}{\dotfill}\textcolor{darkblue}{//\,\textrm{#1}}}

\makeatletter
\@ifundefined{IEEEPARstart}{}{}%
\@ifundefined{IEEEkeywords}{\newenvironment{IEEEkeywords}{}{} }{}%
\makeatother

\begin{document}

\title{Real-Time Human-Adaptive Task Allocation for Multi-Human Multi-Robot Supervision}

\author{Seabin Lee$^{1\dagger}$, Sujeong Park$^{1\dagger}$, Nayoung Kim$^1$, Sungjin Park$^2$, 
        Haechan Jung$^1$, and Changjoo Nam$^{1*}$%
\thanks{This work has been submitted to the IEEE for possible publication. Copyright may be transferred without notice, after which this version may no longer be accessible.}
\thanks{This work was supported by the National Research Foundation of Korea (NRF) grants funded by the Korea government (MSIT) (RS-2024-00461409 and RS-2024-00411007).

Approval of all ethical and experimental procedures and protocols was granted by Sogang University Institutional Review Board (IRB) under Application NO. SGUIRB-A-2501-04-1.

All authors are with Sogang University, Seoul, Korea. $^1$Dept. of Electronic Engineering. $^2$Dept. of Artificial Intelligence.
$^\dagger$Equal contribution (\emph{Co-first authors}). $^*$Corresponding author: {\tt\small cjnam@sogang.ac.kr}}
}

\maketitle

\begin{abstract}
We propose a human-factor-aware method of allocating robot supervision tasks to multiple human operators. In scenarios where multiple operators occasionally teleoperate multiple robots to help the robots overcome difficulties, the allocation of the supervisory control tasks to humans needs to consider the real-time cognitive states of individual operators. However, most existing methods assume fixed supervisory capacity per operator and overlook fluctuations in the human factors such as workload and fatigue. As a result, workload distribution can be unbalanced where some operators become overloaded while the others remain underused. Our method dynamically regulates supervisory capacity and allocates tasks in a way that maintains balanced mental workload, prevents overload, and improves overall team performance. The allocation method uses a greedy strategy that minimizes estimated operator workloads with task prioritization. Robots are assigned to operators by reflecting their current supervisory capacity where the required effort depends on the types of tasks. In the user study, the analysis across predefined time intervals shows that the proposed method consistently achieves higher performance and lower behavioral signs of fatigue compared to a baseline method that does not consider human factors. These results highlight adaptive capacity adjustment as an effective preventive mechanism for sustaining operator performance in long-duration, high-demand settings.
\end{abstract}

\begin{IEEEkeywords}
Human-robot interaction, multi-robot systems, supervisory control
\end{IEEEkeywords}

\section{Introduction}
\label{sec:intro}

\begin{figure}
    \makebox[\linewidth][l]{
        \hspace*{0.05\linewidth}
        \includegraphics[width=0.9\linewidth]{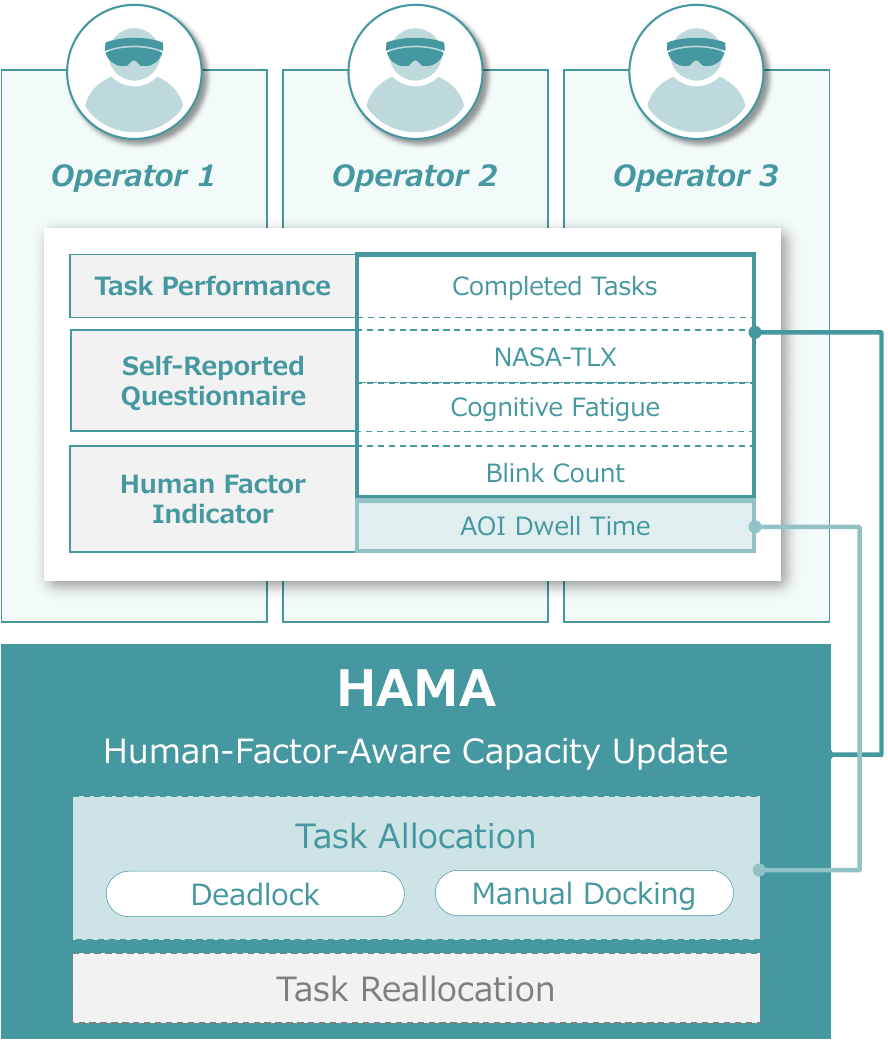}%
    }
    \caption{Overview of HAMA. Inputs include NASA-TLX workload, subjective fatigue, blink count, success count, and AOI dwell time. The system updates operator capacity by predefined time intervals and assigns robots online based on AOI dwell time. Operators perform teleoperation for robots that are in deadlock or help-signal situations. The diagram shows three operators and the flow from sensing to assignment.}
    \label{fig:overview}
\end{figure}

The deployment of multi-robot systems in real-world environments continues to expand, aiming toward full autonomy. Multi-robot task allocation (MRTA) increases throughput and coordination by determining which robot performs which task~\cite{gerkey2004formal}. Although MRTA followed by task and motion planning advances autonomy, current systems still require human supervisors to intervene when robots fail to deal with unexpected events and dynamic changes~\cite{wong2017workload}. Human supervision remains essential because it pairs robotic precision with human judgment and can enhance safety and accuracy. This means that another task allocation is required between human operators and the dynamically emerging supervision tasks. 

However, there are only a few studies on multi-human task allocation in the context of supervising multi-robot systems, and the existing ones typically assume fixed throughput of the human operators, overlooking variations in workload, fatigue, and task performance~\cite{chen2024optimal,meng2024research,cai2022scheduling}. This assumption would no longer be valid if a small number of operators supervise many robots, as human cognitive resources are fundamentally limited~\cite{wong2017workload}. During prolonged supervision, workloads tend to increase and performance can decline, requiring the human interface to adapt to the cognitive and physiological states of operators in human-in-the-loop settings~\cite{kolb2022leveraging}. In addition, operators may differ in skill and performance even for the same type of tasks. Therefore, an effective human-task allocation must not only satisfy mission requirements but also account for real-time human factors and technical capacity. 

We propose \underline{H}uman-factor-\underline{A}ware \underline{M}ulti-robot \underline{A}llocation (HAMA), an adaptive method for multi-human multi-robot (MHMR) supervision that maintains robot task performance while considering human states. As illustrated in Fig.~\ref{fig:overview}, HAMA estimates human factors such as workload and fatigue and updates the supervisory capacity of each operator (i.e., the maximum number of tasks that can be supervised) at predefined time intervals. We focus on two task types that require supervisory intervention: multi-robot deadlock resolution and docking at charging stations, where failures frequently occur in real-world scenarios. For missions with changing task priorities, HAMA supports reallocation to prioritize more urgent robots over earlier assignments. HAMA also accounts for differences in human factors between task types and across operators by using Area-Of-Interest (AOI) dwell time to estimate visual demand. Then robots requiring teleoperation are assigned to operators based on task priority, visual demand, and current capacity.

HAMA introduces dynamic supervisory capacity that adapts to human factors and mission performance, replacing fixed capacity limits used in prior systems. By integrating subjective reports, behavioral indicators, and physiological signals into a unified update rule, HAMA provides individualized operator capacities and enables real-time robot–operator reallocation as middleware. In user studies, HAMA improved team performance by about 10\% and reduced blink counts by 16\% compared to a fixed-capacity baseline, demonstrating that adaptive capacity can increase multi-robot throughput while mitigating fatigue in long-duration, high-demand settings.

\section{Related Work}
\label{sec:relate}

\subsection{Task Allocation Strategies}
The task allocation problem has been widely studied to improve efficiency and mission throughput in multi-robot systems, with more recent work incorporating safety considerations such as collision avoidance to enable scalable and reliable coordination~\cite{gerkey2004formal, chen2022multi,cbs-ta,mrta-rm}. While most MRTA formulations focus on allocating tasks among robots, human-in-the-loop systems additionally require allocating robots to human operators~\cite{zigoris2003balancing, dahiya2022scalable}. Prior studies have shown that the number of robots an operator can manage is closely tied to workload and decision time, exhibiting strong sensitivity to human state~\cite{cummings2007developing}. Building on this insight, collaborative human–robot scheduling approaches explicitly model human assignments, availability, and performance constraints within the optimization process~\cite{gombolay2013fast, lippi2021mixed}. Together, these works indicate that operator capacity should be treated as a variable influenced by human factors, rather than as a fixed constant in the allocation problem.

\subsection{Human-in-the-Loop Control} 
Supervisory control often requires continuous monitoring, managing multiple robots simultaneously, and making rapid decisions in time-sensitive situations. These demands place heavy strain on attention and working memory, which can be depleted during prolonged supervision~\cite{wong2017workload, wu2022task}. As a result, operator workload and fatigue have a direct impact on task performance~\cite{velagapudi2008scaling}. To mitigate these effects, prior studies have explored interface-level support strategies such as guiding visual attention and adjusting interface transparency~\cite{chen2022multi, bahodi2024show}. However, such approaches typically intervene only after performance degradation becomes apparent and do not adjust task allocation in response to ongoing changes in operator state. Moreover, even under identical task conditions, operators differ substantially in how quickly they become fatigued or overwhelmed, and these individual differences can affect overall team performance.

Motivated by these limitations, recent work has begun to integrate human factors directly into task allocation frameworks~\cite{jo2024cognitive, ali2022considerations}. For example, \cite{jo2024cognitive} proposed a model that incorporates cognitive workload and affective state when distributing tasks across multiple operators. However, such methods rely on pre-recorded data and trained models, making it difficult to reflect real-time changes or adapt to individual differences during live operation. Despite these advances, empirical research on task allocation frameworks that dynamically update operator capacity and adapt task distribution in real time remains limited. Accordingly, the present study examines how real-time estimation of operator state can enable adaptive and balanced task distribution in multi-human–multi-robot environments.

\subsection{Human Factors in Measuring Cognitive States}
In practical teleoperation settings, operator state cannot be fully explained by a single factor~\cite{ijtsma2019computational, talebpour2019adaptive}, as cognitive workload and fatigue interactively influence overall performance. 
For example, even under low workload conditions, accumulated cognitive fatigue can lead to performance degradation. In contrast, operators with lower fatigue can maintain performance despite high workload demands. Consequently, relying on a single measure introduces limitations in sensitivity and interpretability. 

To address this limitation, recent studies~\cite{hopko2021effect, jo2024cognitive} have integrated physiological, behavioral, and self-reported questionnaire data to achieve more comprehensive estimation of operator state. Each measure captures a distinct aspect of the human cognitive state~\cite{van2017effects}, and their integration provides a more comprehensive and complementary representation of human performance dynamics~\cite{jo2024cognitive}. 
Reliable task allocation in MHMR systems requires accurate estimation of the cognitive state for each human operator. While many previous studies~\cite{wang2016adaptive, frame2020development} have focused on real-time cognitive workload as a predictor of immediate task demands, prolonged operation also induces cognitive fatigue, which gradually depletes mental resources over time. To maintain consistent supervision performance and prevent overload, it is important to consider both mental workload and fatigue together~\cite{cortes2025attention}.

\begin{figure*}[!h]
    \centering
    \includegraphics[width=0.99\textwidth]{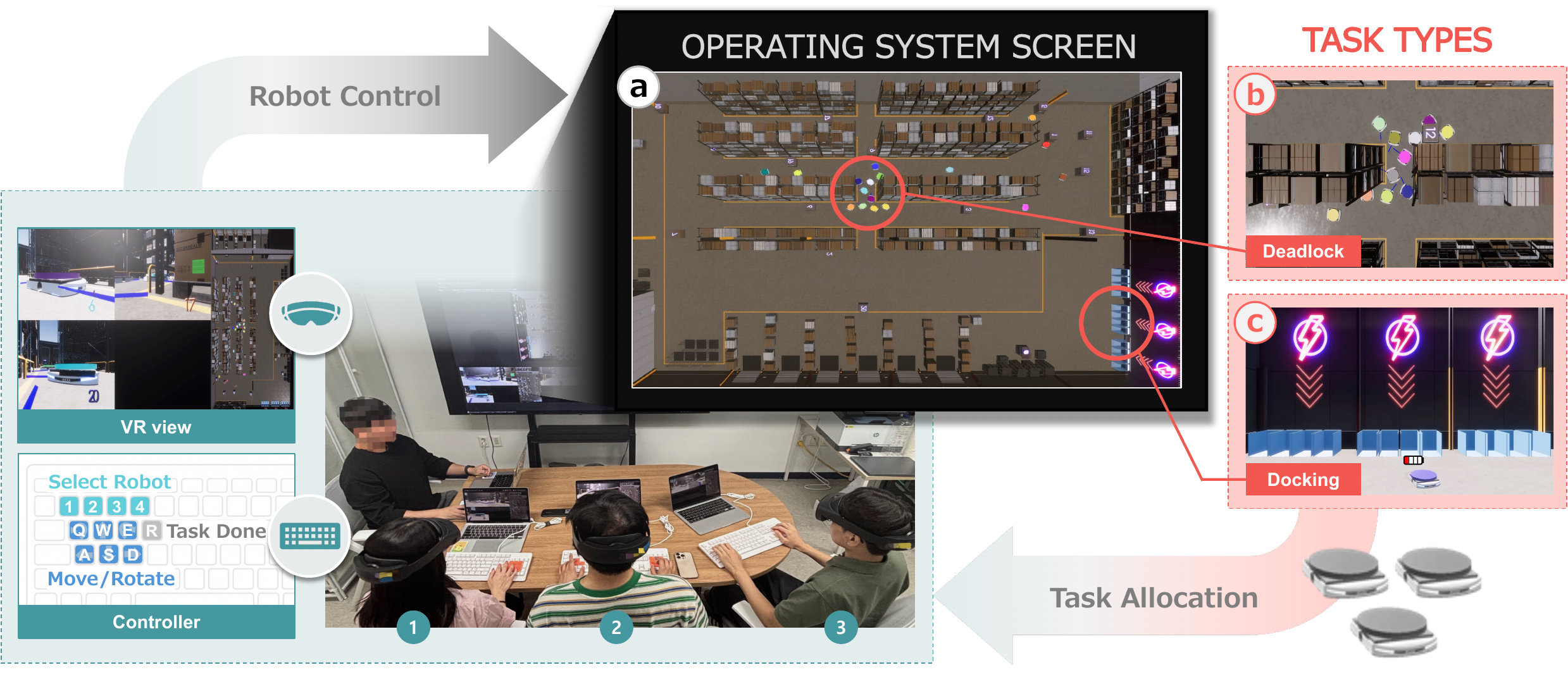}
    \caption{Overview of the experimental setup. (a) Full view of the simulated warehouse environment. Multiple robots navigate passages to perform tasks such as deadlock resolution and charging station docking. (b) Example of a deadlock situation involving multiple robots in a narrow passage. (c) Docking zone with charging indicators. Robots must request help to align correctly for docking. Keyboard control scheme is shown at the bottom-left, and the operator display at the bottom-center.}
    \label{fig:full-exp}
\end{figure*}

\subsubsection{Mental Workload}
Mental workload represents the cognitive effort required during task execution and can fluctuate based on task complexity, frequency of attention shifts, and decision-making demands~\cite{cain2007review, van2018understanding}. As noted by Wickens~\cite{wickens2002multiple}, excessive load on shared cognitive channels can lead to interference and reduce task efficiency. Although NASA-TLX~\cite{hart2006nasa} is widely used to measure workload subjectively, it lacks the temporal resolution to track changes continuously during operations.

To overcome this limitation, several studies ~\cite{ahlstrom2021eye, kunze2019automation} have employed objective indicators such as AOI dwell time obtained through eye-tracking. AOI dwell time reflects how long an operator focuses on specific elements of the interface, making it a sensitive real-time indicator of cognitive workload~\cite{cortes2025attention}. These patterns vary by individual and task types, offering insight into the perceived difficulty of tasks. Accordingly, combining subjective workload rating with objective eye-tracking measures can provide a more accurate and robust estimate of mental workload than either modality alone~\cite{naeeri2021multimodal}.

\subsubsection{Cognitive Fatigue}
Cognitive fatigue refers to the gradual decline in cognitive efficiency that occurs during prolonged task engagement~\cite{wallace2003predicting, borragan2017cognitive}. It results from the depletion of attentional and mental resources over time, leading to reduced alertness, attention span, and slower decision-making. According to the vigilance decrement model~\cite{parasuraman1979memory}, sustained cognitive effort without sufficient recovery can diminish performance and reduce operational reliability. In remote environments such as teleoperation~\cite{lee2022challenges}, supervisory control of autonomous systems~\cite{chen2010supervisory}, or prolonged VR-based monitoring~\cite{wei2021multi}, fatigue can gradually accumulate over time. This can undermine human-system coordination and increase the risk of delayed or incorrect responses.

While subjective fatigue scales can capture perceived tiredness, they are often administered periodically and cannot reflect continuous fluctuations during operations. To complement self-reported measures, several studies employed physiological indicators such as blink rate, which has been shown to increase with cognitive fatigue~\cite{dai2023detection, sun2022investigation} and reduced engagement~\cite{maffei2019spontaneous}. A higher blink rate typically reflects increased cognitive fatigue, serving as a sensitive indicator of resource depletion over time~\cite{dai2023detection}. The blink rate is non-intrusive, continuously measurable, and effective in detecting fatigue in real time without interfering with task performance~\cite{frame2020development}. This makes it particularly suitable for supporting task allocation based on real-time fluctuations in cognitive state.

\section{Problem Definition}
\label{sec:problem}
We consider a warehouse scenario in which multiple mobile robots navigate narrow passages formed by densely packed shelving units (as shown in Fig.~\ref{fig:full-exp}a). A team of human operators remotely supervise the robots via a monitoring interface. While navigation is largely autonomous, certain events require human intervention, and the system must allocate operators to robots under limited attention. We focus on two representative intervention-triggering states: \textit{deadlock}, where the moves of a robot are constrained by surrounding robots or obstacles in narrow spaces (Fig.~\ref{fig:full-exp}b), and \textit{manual docking}, which occurs when autonomous docking fails near battery depletion owing to state estimation uncertainty (Fig.~\ref{fig:full-exp}c). We note that docking tasks occur in proximity to the designated docking stations. Although illustrated in a logistics setting, our approach applies broadly to multi-robot systems in which human intervention can efficiently resolve challenging situations.

Let $\mathcal{R}$ denote the set of robots and $\mathcal{O}$ denote the set of operators. Each operator $o \in \mathcal{O}$ can supervise at most $C_o \ge 1$ robots concurrently, whose upper bound is determined by the cognitive capability of the operator. As cognitive capability can fluctuate, the system regularly updates $C_o$ at predefined time intervals, called \textit{sessions}. Since the operator supervises the robots through a user interface, $C_o$ cannot exceed the number of monitoring windows where each window transmits the ego-view of a robot. In the running example shown in Fig.~\ref{fig:full-exp}, a VR display with four monitoring windows is used. Each monitoring window displays a single supervised robot, and the operator selects which robot to intervene for among the presented options via keyboard input. In our work, the VR display is chosen to provide an immersive environment as well as measure human state indicators, though the interface is not limited to VR\footnote{For example, a computer monitor with an eye-tracker can be another choice.}. The system maintains a \textit{supervision task queue} $\mathcal{Q} \subseteq \mathcal{R}$ that contains robots currently requiring human supervision (i.e., deadlock or manual docking states). The queue is updated continuously as events occur and are resolved. Our objective is to assign robots in $\mathcal{Q}$ to operators in $\mathcal{O}$ to reduce the mental workload of the operators while satisfying the capacity constraints (i.e., $C_o$ for all $o \in \mathcal{O}$).

\section{HAMA: Human-factor-Aware Multi-robot Allocation}
\label{sec:alg}
We propose an algorithm that computes a human-factor–aware task allocation using the capacity of operators estimated based on real-time human state indicators (i.e., mental workload, cognitive fatigue, and blink counts). It consists of (i) the method for updating the operator capacity from real-time measurements and (ii) the allocation method based on the capacity estimates. 

\subsection{Human-factor-aware Capacity Update}
\label{sec:capacity}
Our method measures the initial capacity based on cognitive ability and updates it periodically using the real-time human state indicators as described in Alg.~\ref{alg:capa_update}. The initial capacity is determined from the measured cognitive ability of each operator through 2-back tests~\cite{owen2005n, zhu2024age} before the experiment begins. Three test results of each operator are summed into a cognitive ability score for improved measurement reliability. The capacities are computed proportionally to the cognitive ability scores of the operators such that the resulting capacities do not exceed the system limit defined by the number of monitoring windows.\footnote{As an example, we suppose two operators and four monitoring windows so the number of assigned robots cannot exceed four. If two operators have the scores of 70 and 140, their capacities are 2 and 4, respectively.}

After computing the initial capacities, we update them at each session. Let $w_o$ and $f_o$ denote the NASA-TLX score and the subjective cognitive fatigue score of operator $o$ measured in the current session, respectively. These are combined into a single workload indicator $q_o = w_o + f_o$ (line~\ref{line:qo}). In addition, let $u_o$ denote the number of successfully completed tasks and $b_o$ the number of eye blinks observed during the session. We then compute the relative changes $\Delta q_o$, $\Delta u_o$, and $\Delta b_o$ with respect to the previous session values ($q_o^{\mathrm{prev}}$, $u_o^{\mathrm{prev}}$, and $b_o^{\mathrm{prev}}$), expressed as percentages (lines~\ref{line:delta_start}--\ref{line:delta_end}). When a previous value is zero, the corresponding change is set to zero. Using these terms, we compute
\[
s_{n,o} = -\Delta q_o + \Delta u_o - \Delta b_o
\]
in line~\ref{line:sum}, which determines whether the supervision capacity of operator $o$ should be incremented or decremented. If $\Delta s_{n,o} = s_{n,o} - s_{n,o}^{\mathrm{prev}} > 0$, the system increases $C_o$ by one up to the number of monitoring windows. If $\Delta s_{n,o} < 0$, the system decreases $C_o$ by one down to one. Since the quantities used to compute $\Delta s_{n,o}$ can vary with sensing and system dynamics and may reflect individual characteristics, we rely only on the direction of change (increase or decrease) in the score when updating capacity. Assuming capacity evolves gradually, not rapidly, we apply fixed unit increments or decrements to balance responsiveness and stability (lines~\ref{line:cap_adj_start}--\ref{line:cap_adj_end}). This approach captures improvement or deterioration without overreacting to signal magnitude. After the update, the system stores the current score and session statistics as the next previous values (lines~\ref{line:update_score_start}--\ref{line:update_score_end}).

\begin{algorithm}[t]
\caption{Periodic Capacity Update}
\label{alg:capa_update}
\begin{algorithmic}[1]
{\small
\floatname{algorithm}{Procedure}
\renewcommand{\algorithmicrequire}{\textbf{Input:} }
\renewcommand{\algorithmicensure}{\textbf{Output: }}

\REQUIRE Current capacities $\{C_o\}$, window limit $W_{\max}$, \\ 
$q_o^{\mathrm{prev}}, u_o^{\mathrm{prev}}, b_o^{\mathrm{prev}}, s_{n,o}^{\mathrm{prev}}$, \codecomment{previous scores} \\
$w_o$, $f_o$, $u_o$, $b_o$ \codecomment{recorded per session}
\ENSURE Updated capacities $\{C_o\}$ and updated previous scores

\FORALL{operators $o \in \mathcal{O}$ at each session boundary}
  \STATE $q_o \gets w_o + f_o$ \label{line:qo}
  \STATE $\Delta q_o \gets \mathrm{PercentChange}(q_o, q_o^{\mathrm{prev}})$ \label{line:delta_start}
  \STATE $\Delta u_o \gets \mathrm{PercentChange}(u_o, u_o^{\mathrm{prev}})$
  \STATE $\Delta b_o \gets \mathrm{PercentChange}(b_o, b_o^{\mathrm{prev}})$ \label{line:delta_end}
  \STATE $s_{n,o} \gets -\Delta q_o + \Delta u_o - \Delta b_o$ \label{line:sum}
  \IF{$s_{n,o} > s_{n,o}^{\mathrm{prev}}$} \label{line:cap_adj_start}
    \STATE $C_o \gets \min(C_o + 1,\, W_{\max})$
  \ELSIF{$s_{n,o} < s_{n,o}^{\mathrm{prev}}$}
    \STATE $C_o \gets \max(C_o - 1,\, 1)$ \label{line:cap_adj_end}
  \ENDIF
  \STATE $s_{n,o}^{\mathrm{prev}} \gets s_{n,o}$ \label{line:update_score_start}
  \STATE $q_o^{\mathrm{prev}} \gets q_o$, $u_o^{\mathrm{prev}} \gets u_o$, $b_o^{\mathrm{prev}} \gets b_o$ \label{line:update_score_end}
\ENDFOR
\RETURN $\{C_o\}$
}
\end{algorithmic}
\end{algorithm}

\subsection{Task Allocation Strategy}
\label{sec:allocation}
HAMA allocates supervision requests greedily in an online manner whenever the priority queue $Q$ is non-empty and at least one available operator is assigned fewer tasks than their current capacity. It assigns the task at the head of $Q$ to the operator whose relative workload would be lowest after the assignment. The relative workload does not need to be measured on an absolute scale, as the allocation only requires identifying the operator with the lowest workload among all operators.

The tasks in $Q$ are ordered in descending order of the priority score which is calculated based on task type and urgency. A specific rule for scoring tasks can differ from the mission environments and task types. In our logistics scenario, the maximum possible score of the docking task is higher than the deadlock task as delayed intervention can lead to critical battery depletion. Since all docking tasks occur in proximity to the docking stations, we do not differentiate the importance among them. The scores of deadlock tasks are calculated based on the distance from the goal locations. Robots that are farther from their goal are handled earlier. This strategy encourages robots to be distributed more widely, thereby securing local free space around the deadlock and reducing the likelihood of secondary deadlocks.

On the other hand, the workload of an operator is determined by considering the difficulty of the assigned tasks. In human supervision, task difficulty is perceived subjectively and can vary across operators. For instance, one operator may find deadlock resolution more demanding, whereas another may perceive manual docking as more challenging. Because workload itself is inherently subjective, such individual differences should be reflected in its estimation.

To account for this variability, we estimate the relative difficulty of different task types for each operator using their average AOI dwell times measured in the previous session. Specifically, we compute the ratio of average dwell times between task types and use this ratio to scale the base unit workload assigned to each task. For example, if an operator exhibited average AOI dwell times of $30$ seconds for docking tasks and $45$ seconds for deadlock tasks in the previous session, a ratio of $1:1.5$ is applied in the next session. As a result, deadlock tasks are assigned a workload weight of $1.5$, while docking tasks retain a unit workload.

\begin{figure}[t]
    \centering
    \includegraphics[width=\linewidth]{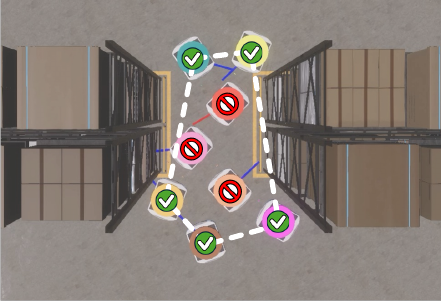}
    \caption{Example of a deadlock cluster in a narrow warehouse passage. Green check marks indicate robots identified as boundary robots, based on convex hull location and directional clearance. Robots marked with the red alert icon are internal to the deadlock cluster and cannot move until boundary robots are cleared.}
    \label{fig:boundary_robot}
\end{figure}

In the case of deadlock resolution, precedence constraints must be enforced, whereby robots outside a deadlock cluster must first move away to enable inner robots to escape. As the composition of the deadlock cluster evolves during deadlock resolution, the set of robots to be prioritized must be updated dynamically. We identify such outer robots by computing the convex hull of the robots in the deadlock cluster and designating those on the hull boundary as outer robots, as illustrated in Fig.~\ref{fig:boundary_robot}. For operators assigned with inner robots during deadlock resolution, HAMA replaces those inner robots with the currently identified outer robots.

\section{User Study}
\label{sec:user_study}
We run a user study to examine the effectiveness of our allocation method considering human factors in terms of the task performance, cognitive workload, and fatigue of the operators. 

\subsection{Participants}
\label{sec:participants}
The study involved 30 participants, grouped into 10 teams of three operators each. The participants included 23 males and 7 females, with an average age of 26.4 years (SD = 3.23). All participants were between 20 and 35 years of age and reported no issues with wearing or interacting with VR devices. None had prior experience with the proposed allocation system. This study received approval from the Institutional Review Board (IRB) of Sogang University (SGUIRB-A-2501-04-1). The participants received $\$14$ as compensation.

\subsection{Apparatus}
\label{sec:Apparatus}
The equipment used in this experiment included Meta Quest Pro (VR device), and a keyboard used as a controller. The VR device has a resolution of 1800$\times$1920 pixels per eye and a maximum field of view of $106^\circ$. Eye tracking data were recorded using the built-in tracker of the device at the default setting. The vendor does not provide a validated accuracy specification for this setting. We therefore ran a standard calibration before the experiment and focused on relative measures such as blink count and AOI dwell time rather than absolute gaze error. All equipment was connected to a desktop computer with an AMD 9950X CPU, 64GB RAM, and an NVIDIA GeForce RTX 5080 GPU. The simulation was built in the Unity Engine using a publicly available industrial asset\footnote{\url{https://assetstore.unity.com/packages/3d/environments/industrial/unity-warehouse-276394}}.

\begin{figure}[t]
    \centering
    \includegraphics[width=\linewidth]{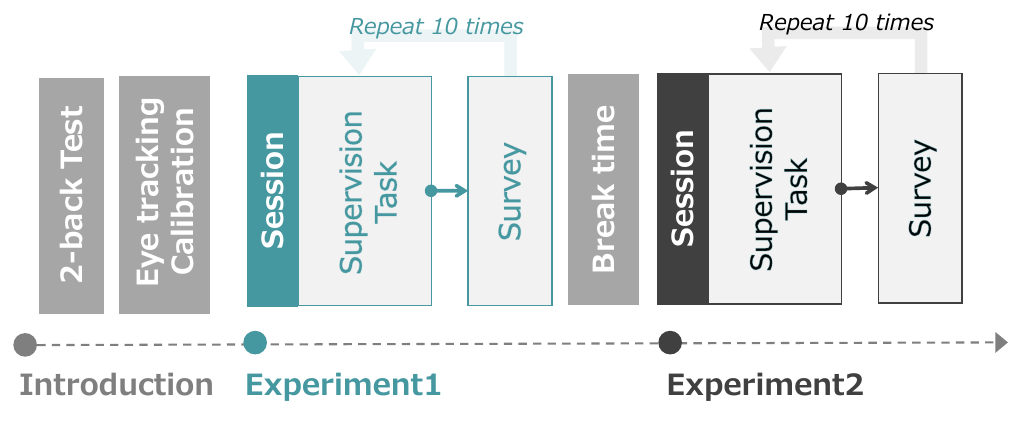}
    \caption{Study timeline. After briefing, 2-back tests, and eye-tracker calibration, teams completed two allocation conditions in counterbalanced order. Each condition consisted of 10 repetitions of a 4-minute supervision task followed by a brief survey. A scheduled break separated the two blocks. Total time per team was about 90 min ($\approx$80 min task time).}
    \label{fig:exp_procedure}
\end{figure}

\subsection{Procedure}
\label{sec:Procedure}
An overview of the experiment procedure is shown in Fig.~\ref{fig:exp_procedure}. Before the experiment, participants were briefed on the study objectives and procedures and provided informed consent. Each team with three participants is informed that they are going to supervise multiple robots in a shared warehouse workspace where they work together on two types of supervision tasks, which are deadlock resolution and manual docking. After that, each participant completed the 2-back cognitive test three times to determine the initial operator capacity as explained in Sec.~\ref{sec:capacity}. We then checked eye-tracking calibration to ensure stable gaze recording. At the end of each session, participants completed NASA-TLX and a subjective fatigue rating. The next session started after all team members submitted their responses.

Each team completed ten 4-minute sessions for each of the two methods (the proposed method and a baseline), for a total of 20 sessions. Repeating each condition ten times allows performance trends to be observed while providing sufficient adaptation time, consistent with prior work involving newly generated task configurations and randomized robot trajectories~\cite{hopko2021effect}. The entire experiment lasted approximately 90 minutes per team, including 80 minutes of task execution and scheduled breaks. 

\subsection{Experimental Design}
\label{sec:Experimental_design} 

We employed a within-subjects design in which each team experienced both allocation conditions: a baseline strategy and the proposed human-factor-aware method. To mitigate learning effects, the order of conditions was counterbalanced across teams. 

\begin{figure}[t]
    \centering
    \includegraphics[width=0.47\textwidth]{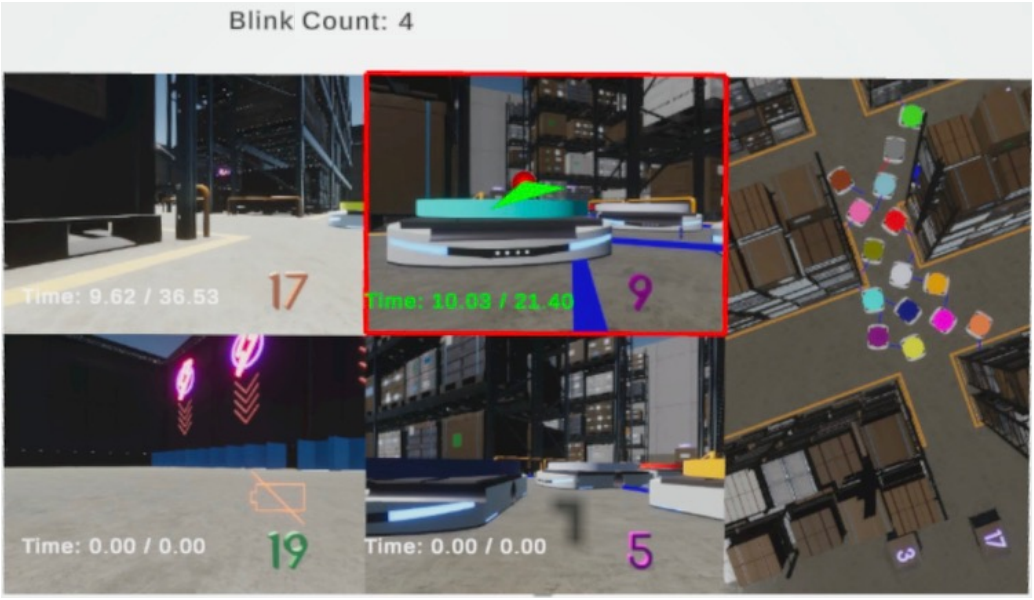}
    \caption{View of the VR control interface as seen by an operator. Multiple robot camera feeds are shown. AOI dwell time and gaze rays were only visible during debugging, not in the actual experiment.}
    \label{fig:vr_interface}
\end{figure}

Operators supervised 20 mobile robots in a warehouse through the VR display and a physical keyboard. Each monitoring window in the VR display shows a single supervision task where the operators could choose which robot to intervene using the keyboard (the key mapping is shown in the bottom-left of Fig.~\ref{fig:full-exp}). Fig.~\ref{fig:vr_interface} shows the four ego-view monitoring windows in the VR display. The number of monitoring windows is determined based on prior studies~\cite{trouvain2002evaluation, rey2024many}, which indicate that remote operators can effectively supervise up to four robots. A top-view window on the side summarizes the overall warehouse state.

We compared HAMA against a non-adaptive baseline in the same task environment, which maintained a fixed operator capacity of four. Additionally, attention-based load estimation was disabled by applying equal weighting to all task types. However, boundary-based reallocation remained enabled. This setting was designed to isolate the effects of capacity update and attention-based weighting. Table~\ref{tab:conditions} summarizes which parts of the allocation framework were enabled in each condition.

\begin{table}[t]
    \caption{Allocation features in HAMA and baseline conditions}
    \label{tab:conditions}
    \centering
    \begin{tabular}{lcc}
        \hline
        Condition                      & HAMA        & Baseline    \\
        \hline
        Capacity update                & \checkmark  & $\times$    \\
        Attention-based weighting & \checkmark  & $\times$    \\
        Boundary-based reallocation    & \checkmark  & \checkmark  \\
        \hline
    \end{tabular}
\end{table}

\subsection{Measures}
\label{sec:Measures_and_Analysis}
We measured self-reported questionnaires, human indicators (e.g., blink count and AOI dwell time), and the number of successfully completed tasks to evaluate human state and performance. 

\subsubsection{Self-reported Questionnaire}
The self-reported measures of human state (per session) include:

\begin{itemize}
    \item Mental Workload: Measured by NASA-TLX, which evaluates six dimensions (mental demand, physical demand, temporal demand, perceived performance, effort, and frustration), rated on a 10-point Likert scale
    \item Cognitive Fatigue Rating: Measured using a single-item rating about fatigue ``What is your level of fatigue?'' rated on a 10-point Likert  (1 = low, 10 = high)
\end{itemize}

\subsubsection{Human Factor Indicator}
Behavioral and physiological data were obtained from the VR eye-tracker (in real-time), which include:

\begin{itemize}
    \item Blink Count: The number of eye blinks per session where a higher blink rate can indicate fatigue or reduced vigilance during prolonged supervision~\cite{dai2023detection}
    \item AOI Dwell Time: The duration of gaze fixation within the ego-view and the top-view window for the robot currently supervised by the operator, which was used to estimate operator-specific relative task difficulty for task-type weighting in the allocation algorithm
\end{itemize}

\subsubsection{Task Performance}
As a task performance per operator, the number of tasks successfully completed were recorded at the end of each session.

\subsection{Analysis}
\label{sec:Data_collection_and_Analysis}
Statistical analyses were performed using IBM SPSS Statistics with a significance level of $0.05$. A two-way repeated-measures ANOVA (RM-ANOVA) was conducted to evaluate the effects of the algorithm and session. When significant main effects were observed, post-hoc Bonferroni-adjusted comparisons were conducted to assess differences between conditions. No significant interaction was found between two factors $(p > 0.05$). During the ten experimental sessions, HAMA progressively adjusted the capacity of each operator using the human factors. This process reflected the cumulative influence of adaptation over time. Because the adaptive algorithm requires data from the previous session to update operator capacity, adaptation begins only after session two. Accordingly, we conducted RM-ANOVA using data from sessions 3 through 10.

\section{Results}
\label{sec:results}

\begin{table*}[h]
\centering
\caption{Session-wise mean $\pm$ SD (Sessions 3–10).
Higher is better for Performance ($\uparrow$), and lower is better for the other measures ($\downarrow$).}
\label{tab:all_met}
\resizebox{\textwidth}{!}{%
\begin{tabular}{ll*{9}{c}}
\toprule
\textbf{Metric} & \textbf{Method} & \multicolumn{9}{c}{\textbf{Session}} \\
\cmidrule(lr){3-11}
& & \textbf{3} & \textbf{4} & \textbf{5} & \textbf{6} & \textbf{7} & \textbf{8} & \textbf{9} & \textbf{10} & \textbf{Overall} \\
\midrule
\multirow{2}{*}{NASA-TLX $\downarrow$}          & HAMA     & 29.03 ± 7.65  & 29.73 ± 7.73  & 30.43 ± 8.85  & 30.80 ± 9.54  & 32.30 ± 8.78  & 31.60 ± 9.03  & 31.57 ± 9.54  & 32.00 ± 9.31  & 30.93 ± 8.77  \\
                                   & Baseline        & 30.03 ± 7.96  & 30.93 ± 8.10  & 30.53 ± 8.52  & 30.77 ± 8.89  & 31.17 ± 8.96  & 31.03 ± 9.60  & 30.87 ± 9.80  & 31.87 ± 10.16 & 30.90 ± 8.91  \\
\midrule
\multirow{2}{*}{Cognitive Fatigue $\downarrow$} & HAMA     & 4.07 ± 2.02   & 4.57 ± 2.03   & 4.73 ± 2.16   & 5.17 ± 2.38   & 5.67 ± 2.50   & 5.67 ± 2.47   & 5.73 ± 2.43   & 5.57 ± 2.33   & 5.15 ± 2.34   \\
                                   & Baseline        & 4.53 ± 2.18   & 4.90 ± 2.20   & 5.20 ± 2.44   & 5.53 ± 2.53   & 5.90 ± 2.59   & 5.93 ± 2.60   & 6.10 ± 2.38   & 6.23 ± 2.51   & 5.54 ± 2.47   \\
\midrule
\multirow{2}{*}{Blink Count $\downarrow$}          & HAMA     & 44.47 ± 42.33 & 49.30 ± 46.65 & 53.90 ± 47.63 & 58.47 ± 56.60 & 58.90 ± 58.75 & 62.47 ± 53.67 & 58.80 ± 56.87 & 61.03 ± 53.62 & 55.92 ± 51.86  \\
                                   & Baseline        & 59.83 ± 47.12 & 57.07 ± 48.88 & 64.50 ± 56.93 & 62.30 ± 48.08 & 67.30 ± 57.59 & 75.37 ± 62.82 & 76.23 ± 52.37 & 68.50 ± 47.99 & 66.39 ± 52.61  \\
\midrule
\multirow{2}{*}{Performance $\uparrow$}       & HAMA     & 17.27 ± 7.67  & 19.90 ± 8.19  & 18.43 ± 7.16  & 16.63 ± 5.99  & 18.43 ± 7.02  & 16.73 ± 6.91  & 17.77 ± 7.61  & 17.73 ± 5.85  & 17.86 ± 7.06  \\
                                   & Baseline        & 16.07 ± 5.91  & 17.67 ± 5.72  & 16.57 ± 7.77  & 17.33 ± 5.25  & 15.73 ± 4.72  & 15.30 ± 5.77  & 15.40 ± 5.86  & 15.67 ± 5.55  & 16.22 ± 5.85  \\
\midrule
\multirow{1}{*}{Capacity}       & HAMA     & 2.73 ± 0.74  & 2.70 ± 0.75  & 2.97 ± 0.67  & 2.90 ± 0.80  & 2.90 ± 0.76  & 3.03 ± 0.81  & 3.07 ± 0.69  & 2.93 ± 0.87  & 2.90 ± 0.76  \\
\bottomrule
\end{tabular}
}
\end{table*}

\begin{table}[!t]
{\centering
\caption{Results of two-way RM-ANOVA in experiment.}
\label{tab:anova_results}
\begin{tabular}{llccc}
\toprule
\textbf{Metric} & \textbf{Effect} & \textbf{$F$} & \textbf{$\eta^2_p$} & $post$-$hoc$ \\
\midrule
\multirow{2}{*}{NASA-TLX}          & Session         & 3.342$^{*}$   & 0.504             & 3\textless{}7,10                                                                             \\
                                   & Algorithm       & 0.001         & 0                 & -                                                                                            \\
\midrule
\multirow{2}{*}{Cognitive Fatigue} & Session         & 8.182$^{***}$ & 0.713             & \begin{tabular}[c]{@{}c@{}}4,5\textless{}6\\ 3\textless{}4,5\textless{}7,8,9,10\end{tabular} \\
                                   & Algorithm       & 1.546         & 0.051             & -                                                                                            \\
\midrule
\multirow{2}{*}{Blink Count}          & Session         & 1.558         & 0.322             & -                                                                                            \\
                                   & Algorithm       & 6.077$^{*}$   & 0.173             & -                                                                                            \\
\midrule
\multirow{2}{*}{Performance}       & Session         & 2.541$^{*}$   & 0.436             & 8\textless{}4                                                                                \\
                                   & Algorithm       & 4.362$^{*}$   & 0.131             & -                                                                                            \\
\bottomrule
\end{tabular}}

\vspace{0.5em}
\footnotesize
$^{*}p < 0.05$, $^{**}p < 0.01$, $^{***}p < 0.001$.
\end{table}

\begin{figure}[!h]
    \centering
    \includegraphics[width=0.915\linewidth]{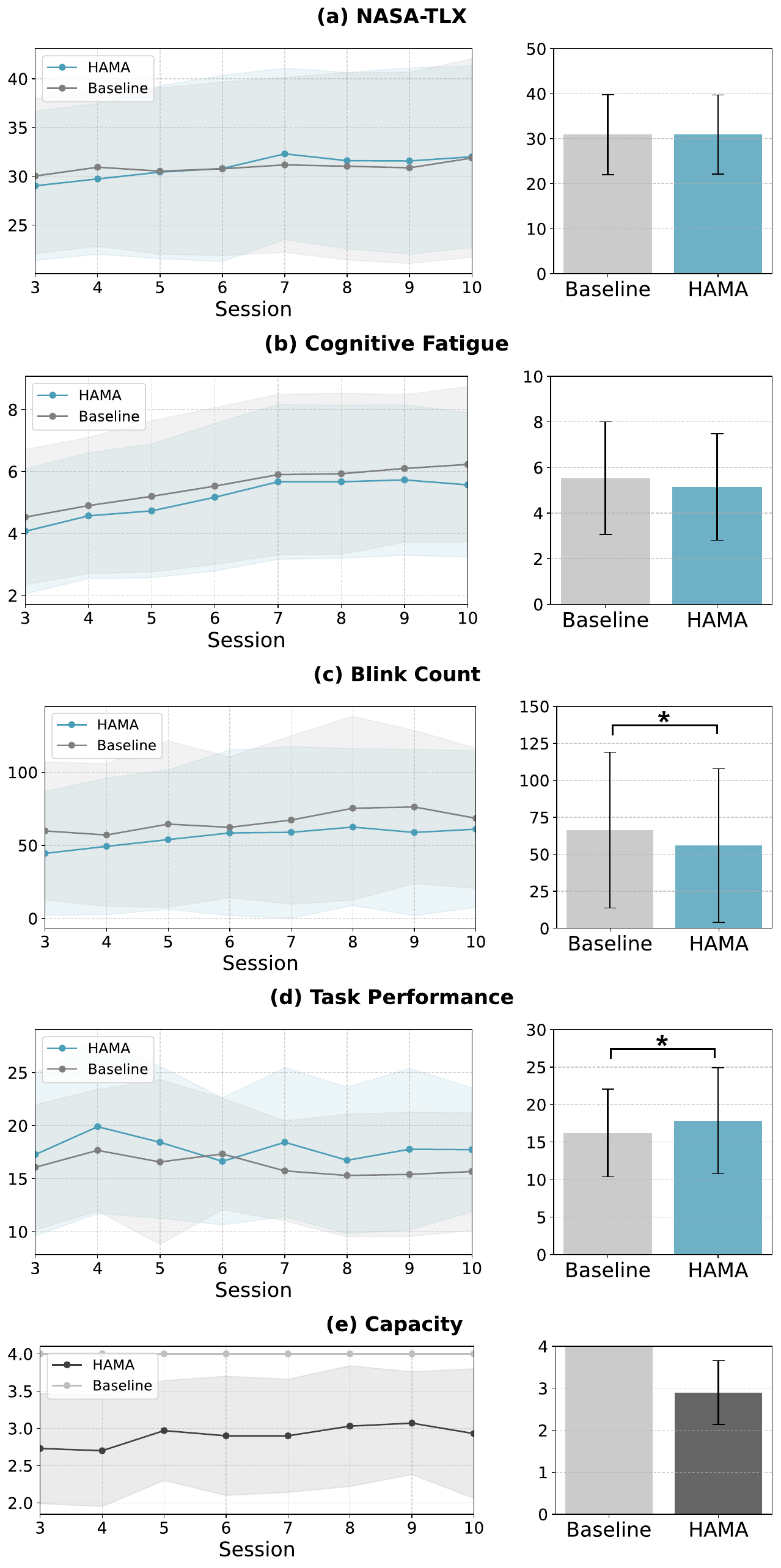}
    \caption{Session-wise trends (left) and overall mean with SD (right) for NASA-TLX, Cognitive Fatigue, Blink Count, Performance, and Capacity. Panels are stacked from top to bottom. Asterisks mark a significant main effect of the algorithm for Blink Count and Performance ($p<0.05$).}
    \label{fig:met}
\end{figure}

This section presents the results of the experiment, comparing cognitive measures and task performance between the baseline and the proposed HAMA algorithm. The details of the RM-ANOVA analysis are presented in Table~\ref{tab:anova_results}. Session-wise means with between-participant standard deviation (SD) for each dependent variable are shown in Table~\ref{tab:all_met}. The Overall column of Table~\ref{tab:all_met} reports the mean and SD of the averages at the participant-level during sessions 3--10. Fig.~\ref{fig:met} shows the session-wise trends and the overall analysis.

\subsection{Self-reported Questionnaire}
For NASA-TLX (Fig.~\ref{fig:met}a), there was no significant effect of algorithm [$F$(1, 29) = 0.001, $p$ = 0.975, $\eta^2_p$ = 0.000]. However, a significant main effect of session was observed [$F$(7, 23) = 3.342, $p$ = 0.013, $\eta^2_p$ = 0.504]. No significant interaction between algorithm and session was found [$F$(7, 23) = 0.992, $p$ = 0.461, $\eta^2_p$ = 0.232]. 

For cognitive fatigue (Fig.~\ref{fig:met}b), HAMA consistently showed lower ratings than baseline, although the main effect of the algorithm did not reach statistical significance [$F$(1, 29) = 1.546, $p$ = 0.224, $\eta^2_p$ = 0.051]. However, a significant main effect of session was observed [$F$(7, 23) = 8.182, $p$ $<$ 0.001, $\eta^2_p$ = 0.713]. No significant interaction between algorithm and session was found [$F$(7, 23) = 0.294, $p$ = 0.949, $\eta^2_p$ = 0.082].

\subsection{Human Factor Indicators}
The analysis revealed a significant main effect of algorithm [$F$(1, 29) = 6.077, $p$ = 0.020, $\eta^2_p$ = 0.173], with HAMA resulting in significantly fewer blinks as shown in Fig.~\ref{fig:met}c. However, no significant effect of session was found [$F$(7, 23) = 1.558, $p$ = 0.198, $\eta^2_p$ = 0.322], indicating that blink count remained relatively stable over time. No significant interaction between algorithm and session was found [$F$(7, 23) = 0.603, $p$ = 0.748, $\eta^2_p$ = 0.155].

\begin{figure}[!t]
    \makebox[\linewidth][l]{
        \hspace*{0.04\linewidth}
        \includegraphics[width=0.7\linewidth]{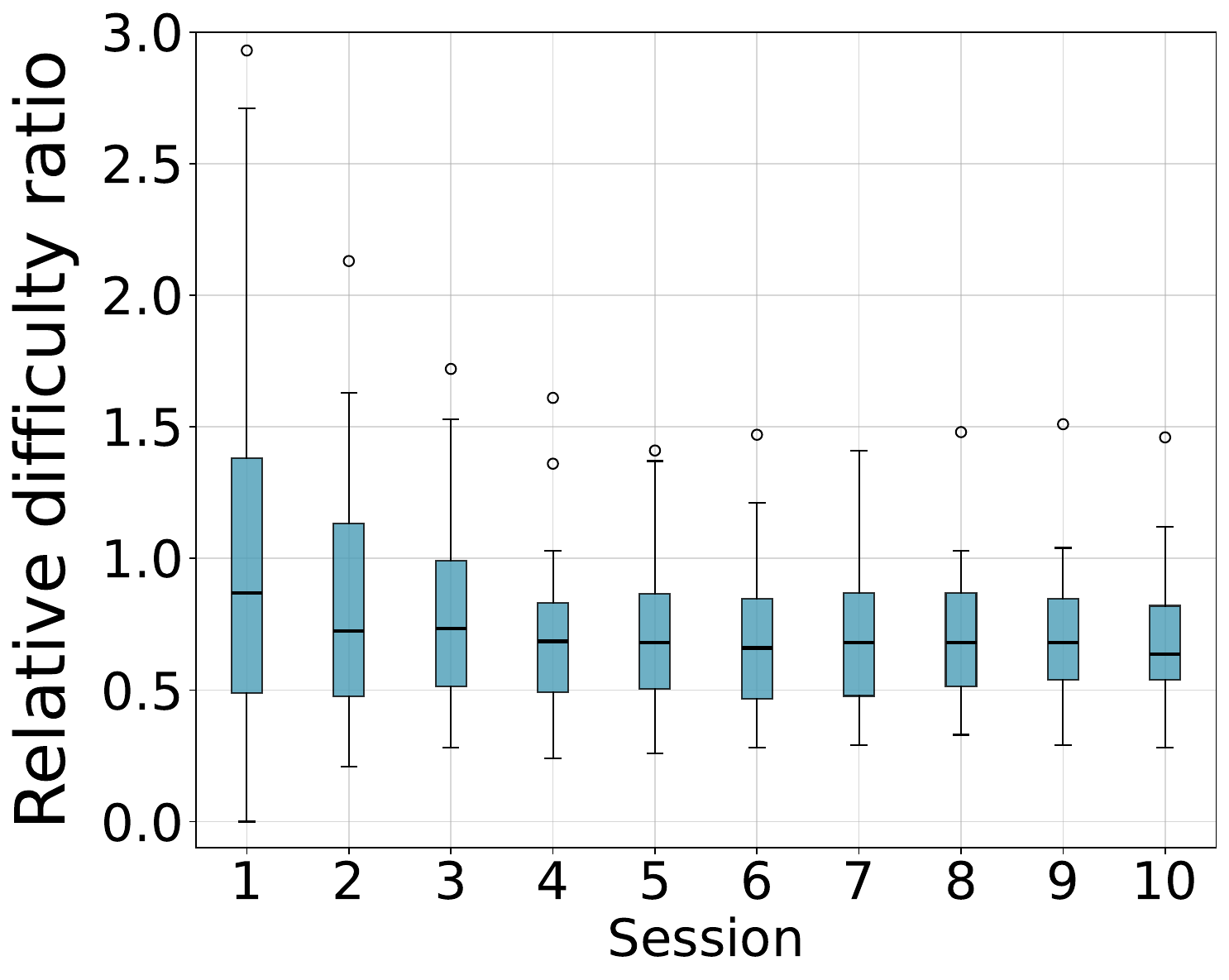}
    }
    \caption{Plot of changes in the average relative difficulty ratio by session.}
    \label{fig:aoi}
\end{figure}

The relative difficulty between the two task types was estimated from AOI dwell time and used internally in HAMA to weight supervision tasks. We define the relative difficulty ratio as the average AOI dwell time during deadlock handling divided by that during manual docking handling. Fig.~\ref{fig:aoi} summarizes this ratio by session. The medians were below 1 across sessions, indicating that manual docking required longer sustained attention than deadlock resolution throughout the experiment.

\subsection{Task Performance}
A significant main effect of the algorithm was found for task performance [$F$(1, 29) = 4.362, $p$ = 0.046, $\eta^2_p$ = 0.131], indicating that HAMA led to a higher number of successfully completed tasks than the baseline (Fig.~\ref{fig:met}d). In addition, there was a significant main effect of session observed [$F$(7, 23) = 2.541, $p$ = 0.043, $\eta^2_p$ = 0.436]. No significant interaction between algorithm and session was found [$F$(7, 23) = 1.421, $p$ = 0.245, $\eta^2_p$ = 0.302].

\subsection{Algorithm}
We also summarize the overall effect of the algorithm on key measures and computational cost. Across sessions 3--10, HAMA increased average task performance from 16.22 to 17.86 (about a 10\% increase) relative to baseline. Over the same sessions, blink count decreased from 66.39 to 55.92 (about a 16\% reduction), indicating lower behavioral signs of fatigue under HAMA. In contrast, NASA-TLX scores changed by less than 1\% between algorithms (30.90 to 30.93), and cognitive fatigue ratings decreased by about 7\% (5.54 to 5.15) without a significant main effect of algorithm.

We also evaluated the computational cost required for allocation decisions. HAMA, despite processing additional human factor data, recorded an average allocation latency of 1.34 ms, compared to 1.26 ms for the baseline. This negligible difference demonstrates that HAMA improves team performance and mitigates physiological fatigue with minimal computational overhead, maintaining real-time responsiveness.

\section{Discussion}
\label{sec:discussion}
We presented HAMA, a human-factor-aware task allocation algorithm for MHMR supervision. We treat supervisory capacity as a controllable variable instead of a fixed constant. The algorithm estimates operator state from behavioral and self-reported subjective indicators to update per-operator capacity at predefined session boundaries and allocates supervision tasks to operators online. As shown in Fig.~\ref{fig:met}e, HAMA adjusted per-operator supervisory capacity across sessions, whereas the baseline remained fixed.

As a result, HAMA improved objective performance while reducing fatigue-related behavior without increasing perceived workload. While HAMA increased the number of successfully completed tasks compared with a fixed-capacity baseline, it reduced blink counts, which indicates lower behavioral signs of fatigue. %

\subsection{Efficiency Gains through Adaptive Allocation}
A key insight is that higher throughput did not require higher individual capacity. In our study, HAMA often assigned fewer concurrent robots per operator than the baseline. Even with this reduction, performance improved. This result suggests that overload can reduce the quality and speed of interventions in MHMR supervision. HAMA mitigates this risk by redistributing supervisory demand toward operators with stable performance signals and away from operators with rising strain signals. The mechanism aligns with the goal of adaptive automation, which aims to keep human demand within a functional range rather than maximizing assigned volume.

\subsection{Behavioral Sensitivity Beyond Subjective Workload Ratings}
The results show a gap between behavioral indicators and subjective reports. The blink count changed with the algorithm, while NASA-TLX and fatigue ratings did not. This gap can occur in high-demand monitoring tasks where perceived workload becomes less sensitive once cognitive resources are saturated. In addition, NASA-TLX and single-item fatigue ratings are collected after each session and may miss short-term fluctuations during the session. These findings support the use of complementary measures when evaluating adaptive allocation systems, since subjective reports alone may not reflect changes in physiological or behavioral state.

\subsection{Limitations}
Several limitations should be considered. The study used a VR interface with a small set of intervention tasks, namely deadlock resolution and manual docking. The participant group had limited prior experience with this supervision setting. In addition, capacity is updated at session boundaries, which prioritizes stable operation and reduces sensitivity to transient fluctuations within a session. If faster adaptation is desirable, a shorter update time can be explored.

\section{Conclusion}
HAMA provides a lightweight and real-time approach for human-adaptive allocation in MHMR supervision. The framework improved team performance and reduced fatigue-related behavior with negligible computational overhead. Future work will test longer deployments and more diverse operator populations. The observed effects across measures open a path toward more principled weighting of the integrated human-factor score, with objective indicators taking a larger role when they show clear algorithm sensitivity and subjective measures serving as complementary signals for robustness.

\bibliographystyle{IEEEtran}
\bibliography{references}

\begin{thebibliography}{10}
\providecommand{\url}[1]{#1}
\csname url@samestyle\endcsname
\providecommand{\newblock}{\relax}
\providecommand{\bibinfo}[2]{#2}
\providecommand{\BIBentrySTDinterwordspacing}{\spaceskip=0pt\relax}
\providecommand{\BIBentryALTinterwordstretchfactor}{4}
\providecommand{\BIBentryALTinterwordspacing}{\spaceskip=\fontdimen2\font plus
\BIBentryALTinterwordstretchfactor\fontdimen3\font minus
  \fontdimen4\font\relax}
\providecommand{\BIBforeignlanguage}[2]{{%
\expandafter\ifx\csname l@#1\endcsname\relax
\typeout{** WARNING: IEEEtran.bst: No hyphenation pattern has been}%
\typeout{** loaded for the language `#1'. Using the pattern for}%
\typeout{** the default language instead.}%
\else
\language=\csname l@#1\endcsname
\fi
#2}}
\providecommand{\BIBdecl}{\relax}
\BIBdecl

\bibitem{gerkey2004formal}
B.~P. Gerkey and M.~J. Matari{\'c}, ``A formal analysis and taxonomy of task
  allocation in multi-robot systems,'' \emph{The International Journal of
  Robotics Research}, vol.~23, no.~9, pp. 939--954, 2004.

\bibitem{wong2017workload}
C.~Y. Wong and G.~Seet, ``Workload, awareness and automation in multiple-robot
  supervision,'' \emph{International Journal of Advanced Robotic Systems},
  vol.~14, no.~3, p. 1729881417710463, 2017.

\bibitem{chen2024optimal}
Q.~Chen and Y.-J. Pan, ``An optimal task planning and agent-aware allocation
  algorithm in collaborative tasks combining with pddl and popf,'' \emph{arXiv
  preprint arXiv:2407.08534}, 2024.

\bibitem{meng2024research}
X.~Meng and Z.~Ding, ``Research on multi-stage path planning for
  human-multi-robot collaboration assembly,'' in \emph{Proceedings of 5th
  International Conference on Artificial Intelligence and Computer
  Engineering}, 2024, pp. 1138--1142.

\bibitem{cai2022scheduling}
Y.~Cai, A.~Dahiya, N.~Wilde, and S.~L. Smith, ``Scheduling operator assistance
  for shared autonomy in multi-robot teams,'' in \emph{Proceedings of the IEEE
  Conference on Decision and Control}, 2022, pp. 3997--4003.

\bibitem{kolb2022leveraging}
J.~Kolb, H.~Ravichandar, and S.~Chernova, ``Leveraging cognitive states in
  human-robot teaming,'' in \emph{Proceedings of the IEEE International
  Conference on Robot and Human Interactive Communication}, 2022, pp. 792--799.

\bibitem{chen2022multi}
S.~Chen, M.~J. O'Brien, F.~Talbot, J.~Williams, B.~Tidd, A.~Pitt, and R.~C.
  Arkin, ``Multi-modal user interface for multi-robot control in underground
  environments,'' in \emph{Proceedings of the IEEE/RSJ International Conference
  on Intelligent Robots and Systems}, 2022, pp. 9995--10\,002.

\bibitem{cbs-ta}
W.~H{\"o}nig, S.~Kiesel, A.~Tinka, J.~W. Durham, and N.~Ayanian,
  ``Conflict-based search with optimal task assignment,'' in \emph{Proceedings
  of the International Conference on Autonomous Agents and Multiagent Systems},
  2018, pp. 757--765.

\bibitem{mrta-rm}
S.~Lee, J.~Sim, and C.~Nam, ``Very large-scale multi-robot task allocation in
  challenging environments via robot redistribution,'' \emph{Robotics and
  Autonomous Systems}, vol. 194, p. 105126, 2025.

\bibitem{zigoris2003balancing}
P.~Zigoris, J.~Siu, O.~Wang, and A.~T. Hayes, ``Balancing automated behavior
  and human control in multi-agent systems: a case study in roboflag,'' in
  \emph{Proceedings of the American Control Conference}, vol.~1, 2003, pp.
  667--671.

\bibitem{dahiya2022scalable}
A.~Dahiya, N.~Akbarzadeh, A.~Mahajan, and S.~L. Smith, ``Scalable operator
  allocation for multirobot assistance: A restless bandit approach,''
  \emph{IEEE Transactions on Control of Network Systems}, vol.~9, no.~3, pp.
  1397--1408, 2022.

\bibitem{cummings2007developing}
M.~L. Cummings and S.~Guerlain, ``Developing operator capacity estimates for
  supervisory control of autonomous vehicles,'' \emph{Human Factors}, vol.~49,
  no.~1, pp. 1--15, 2007.

\bibitem{gombolay2013fast}
M.~Gombolay, R.~Wilcox, and J.~Shah, ``Fast scheduling of multi-robot teams
  with temporospatial constraints,'' \emph{Robotics: Science and Systems},
  2013.

\bibitem{lippi2021mixed}
M.~Lippi and A.~Marino, ``A mixed-integer linear programming formulation for
  human multi-robot task allocation,'' in \emph{Proceedings of the IEEE
  International Conference on Robot and Human Interactive Communication}, 2021,
  pp. 1017--1023.

\bibitem{wu2022task}
H.~Wu, A.~Ghadami, A.~E. Bayrak, J.~M. Smereka, and B.~I. Epureanu, ``Task
  allocation with load management in multi-agent teams,'' in \emph{Proceedings
  of the IEEE International Conference on Robotics and Automation}, 2022, pp.
  8823--8830.

\bibitem{velagapudi2008scaling}
P.~Velagapudi, P.~Scerri, K.~Sycara, H.~Wang, M.~Lewis, and J.~Wang, ``Scaling
  effects in multi-robot control,'' in \emph{Proceedings of the IEEE/RSJ
  International Conference on Intelligent Robots and Systems}, 2008, pp.
  2121--2126.

\bibitem{bahodi2024show}
M.-T. Bahodi, N.~van Berkel, M.~Skov, and T.~Merritt, ``Show me what's wrong:
  Impact of explicit alerts on novice supervisors of a multi-robot monitoring
  system,'' in \emph{Proceedings of the Second International Symposium on
  Trustworthy Autonomous Systems}, 2024, pp. 1--17.

\bibitem{jo2024cognitive}
W.~Jo, R.~Wang, B.~Yang, D.~Foti, M.~Rastgaar, and B.-C. Min, ``Cognitive
  load-based affective workload allocation for multihuman multirobot teams,''
  \emph{IEEE Transactions on Human-Machine Systems}, 2024.

\bibitem{ali2022considerations}
A.~Ali, D.~M. Tilbury, and L.~P. Robert~Jr, ``Considerations for task
  allocation in human-robot teams,'' \emph{arXiv preprint arXiv:2210.03259},
  2022.

\bibitem{ijtsma2019computational}
M.~IJtsma, L.~M. Ma, A.~R. Pritchett, and K.~M. Feigh, ``Computational
  methodology for the allocation of work and interaction in human-robot
  teams,'' \emph{Journal of Cognitive Engineering and Decision Making},
  vol.~13, no.~4, pp. 221--241, 2019.

\bibitem{talebpour2019adaptive}
Z.~Talebpour and A.~Martinoli, ``Adaptive risk-based replanning for human-aware
  multi-robot task allocation with local perception,'' \emph{IEEE Robotics and
  Automation Letters}, vol.~4, no.~4, pp. 3790--3797, 2019.

\bibitem{hopko2021effect}
S.~K. Hopko, R.~Khurana, R.~K. Mehta, and P.~R. Pagilla, ``Effect of cognitive
  fatigue, operator sex, and robot assistance on task performance metrics,
  workload, and situation awareness in human-robot collaboration,'' \emph{IEEE
  Robotics and Automation Letters}, vol.~6, no.~2, pp. 3049--3056, 2021.

\bibitem{van2017effects}
J.~Van~Cutsem, S.~Marcora, K.~De~Pauw, S.~Bailey, R.~Meeusen, and B.~Roelands,
  ``The effects of mental fatigue on physical performance: a systematic
  review,'' \emph{Sports Medicine}, vol.~47, no.~8, pp. 1569--1588, 2017.

\bibitem{wang2016adaptive}
R.~Wang, Y.~Zhang, and L.~Zhang, ``An adaptive neural network approach for
  operator functional state prediction using psychophysiological data,''
  \emph{Integrated Computer-Aided Engineering}, vol.~23, no.~1, pp. 81--97,
  2016.

\bibitem{frame2020development}
M.~E. Frame, A.~S. Boydstun, and J.~S. Lopez, ``Development of an autonomous
  manager for dynamic human-machine task allocation in operational
  surveillance,'' in \emph{Proceedings of the IEEE International Conference on
  Human-Machine Systems}, 2020, pp. 1--4.

\bibitem{cortes2025attention}
C.~J. Cortes, R.~Agbeyibor, J.~Kolb, R.~Bowers, K.~Feigh, and S.~Coogan,
  ``Attention-tunable safety barriers for human-autonomy teaming in specialized
  aviation missions,'' \emph{IEEE Robotics and Automation Letters}, 2025.

\bibitem{cain2007review}
B.~Cain, ``A review of the mental workload literature,'' 2007.

\bibitem{van2018understanding}
B.~B. Van~Acker, D.~D. Parmentier, P.~Vlerick, and J.~Saldien, ``Understanding
  mental workload: from a clarifying concept analysis toward an implementable
  framework,'' \emph{Cognition, technology \& work}, vol.~20, no.~3, pp.
  351--365, 2018.

\bibitem{wickens2002multiple}
C.~D. Wickens, ``Multiple resources and performance prediction,''
  \emph{Theoretical Issues in Ergonomics Science}, vol.~3, no.~2, pp. 159--177,
  2002.

\bibitem{hart2006nasa}
S.~G. Hart, ``{NASA}-task load index ({NASA}-tlx); 20 years later,'' in
  \emph{Proceedings of the human factors and ergonomics society annual
  meeting}, vol.~50, no.~9, 2006, pp. 904--908.

\bibitem{ahlstrom2021eye}
C.~Ahlstr{\"o}m, K.~Kircher, M.~Nystr{\"o}m, and B.~Wolfe, ``Eye tracking in
  driver attention research—how gaze data interpretations influence what we
  learn,'' \emph{Frontiers in Neuroergonomics}, vol.~2, p. 778043, 2021.

\bibitem{kunze2019automation}
A.~Kunze, S.~J. Summerskill, R.~Marshall, and A.~J. Filtness, ``Automation
  transparency: implications of uncertainty communication for human-automation
  interaction and interfaces,'' \emph{Ergonomics}, vol.~62, no.~3, pp.
  345--360, 2019.

\bibitem{naeeri2021multimodal}
S.~Naeeri, Z.~Kang, S.~Mandal, and K.~Kim, ``Multimodal analysis of eye
  movements and fatigue in a simulated glass cockpit environment,''
  \emph{Aerospace}, vol.~8, no.~10, p. 283, 2021.

\bibitem{wallace2003predicting}
J.~C. Wallace, S.~J. Vodanovich, and B.~M. Restino, ``Predicting cognitive
  failures from boredom proneness and daytime sleepiness scores: An
  investigation within military and undergraduate samples,'' \emph{Personality
  and Individual Differences}, vol.~34, no.~4, pp. 635--644, 2003.

\bibitem{borragan2017cognitive}
G.~Borrag{\'a}n, H.~Slama, M.~Bartolomei, and P.~Peigneux, ``Cognitive fatigue:
  A time-based resource-sharing account,'' \emph{Cortex}, vol.~89, pp. 71--84,
  2017.

\bibitem{parasuraman1979memory}
R.~Parasuraman, ``Memory load and event rate control sensitivity decrements in
  sustained attention,'' \emph{Science}, vol. 205, no. 4409, pp. 924--927,
  1979.

\bibitem{lee2022challenges}
J.~S. Lee, Y.~Ham, H.~Park, and J.~Kim, ``Challenges, tasks, and opportunities
  in teleoperation of excavator toward human-in-the-loop construction
  automation,'' \emph{Automation in Construction}, vol. 135, p. 104119, 2022.

\bibitem{chen2010supervisory}
J.~Y. Chen, M.~J. Barnes, and M.~Harper-Sciarini, ``Supervisory control of
  multiple robots: Human-performance issues and user-interface design,''
  \emph{IEEE Transactions on Systems, Man, and Cybernetics, Part C
  (Applications and Reviews)}, vol.~41, no.~4, pp. 435--454, 2010.

\bibitem{wei2021multi}
D.~Wei, B.~Huang, and Q.~Li, ``Multi-view merging for robot teleoperation with
  virtual reality,'' \emph{IEEE Robotics and Automation Letters}, vol.~6,
  no.~4, pp. 8537--8544, 2021.

\bibitem{dai2023detection}
L.~Dai, Y.~Li, and M.~Zhang, ``Detection of operator fatigue in the main
  control room of a nuclear power plant based on eye blink rate, perclos and
  mouse velocity,'' \emph{Applied Sciences}, vol.~13, no.~4, p. 2718, 2023.

\bibitem{sun2022investigation}
L.~Sun, Z.~Guo, X.~Yuan, X.~Wang, C.~Su, J.~Jiang, and X.~Li, ``An
  investigation of the effects of brain fatigue on the sustained attention of
  intelligent coal mine vdt operators,'' \emph{International journal of
  environmental research and public health}, vol.~19, no.~17, p. 11034, 2022.

\bibitem{maffei2019spontaneous}
A.~Maffei and A.~Angrilli, ``Spontaneous blink rate as an index of attention
  and emotion during film clips viewing,'' \emph{Physiology \& Behavior}, vol.
  204, pp. 256--263, 2019.

\bibitem{owen2005n}
A.~M. Owen, K.~M. McMillan, A.~R. Laird, and E.~Bullmore, ``N-back working
  memory paradigm: A meta-analysis of normative functional neuroimaging
  studies,'' \emph{Human brain mapping}, vol.~25, no.~1, pp. 46--59, 2005.

\bibitem{zhu2024age}
S.~Zhu, Q.~Wang, C.~Kan, A.~Geng, Y.~Sui, R.~Zhuang, Y.~Zhu, T.~Wang, L.~Zhu,
  and C.~Guo, ``Age-related cerebral changes during different n-back tasks: a
  functional near-infrared spectroscopy study,'' \emph{Frontiers in aging
  neuroscience}, vol.~16, p. 1437587, 2024.

\bibitem{trouvain2002evaluation}
B.~Trouvain and H.~Wolf, ``Evaluation of multi-robot control and monitoring
  performance,'' in \emph{Proceedings of the IEEE International Conference on
  Robot and Human Interactive Communication}, 2002, pp. 111--116.

\bibitem{rey2024many}
E.~Rey-Becerra and S.~Wischniewski, ``How many robots is too many? findings
  about single-human multiple-robot systems,'' \emph{Human Interaction and
  Emerging Technologies}, vol.~1, no.~1, 2024.

\end{thebibliography}
\end{document}